\documentclass[11pt,a4paper]{article}
\usepackage[hyperref]{naaclhlt2018}
\usepackage{times}
\usepackage{latexsym}
\usepackage{textcomp}
\usepackage{subfig}
\usepackage{pgfplots}
\usepackage{verbatim}
\usepackage{graphicx}  
\usepackage{tabularx}
\usepackage{makecell}
\usepackage{bm}
\usepackage{siunitx}

\usepackage{caption}
\DeclareCaptionFont{10pt}{\fontsize{10pt}{12pt}\selectfont}
\usepackage{url}

\aclfinalcopy 

    \title{An Annotated Corpus for Machine Reading of Instructions \\ in Wet Lab Protocols}

\author{Chaitanya Kulkarni, Wei Xu, Alan Ritter, Raghu Machiraju \\
Department of Computer Science and Engineering\\ 
Ohio State University\\
{\tt \{kulkarni.132,xu.1265,ritter.1492,machiraju.1\}@osu.edu}
}

\date{}

\begin{document}
 \graphicspath{{./images/}}
\maketitle
\begin{abstract}
 We describe an effort to annotate a corpus of natural language instructions consisting of 622 wet lab protocols to facilitate automatic or semi-automatic conversion of protocols into a machine-readable format and benefit biological research. Experimental results demonstrate the utility of our corpus for developing machine learning approaches to shallow semantic parsing of instructional texts. We make our annotated Wet Lab Protocol Corpus available to the research community.\footnote{The dataset is available on the authors' websites.}
\end{abstract}

\section{Introduction}
As the complexity of biological experiments increases, there is a growing need to automate wet laboratory procedures to avoid mistakes due to human error and also to enhance the reproducibility of experimental biological research \cite{king2009automation}.  Several efforts are currently underway to define machine-readable formats for writing wet lab protocols \cite{ananthanarayanan2010biocoder,soldatova2014exact2,vasilev2011software}. The vast majority of today's  protocols, however, are written in natural language with jargon and colloquial 
language constructs that emerge as a byproduct of ad-hoc protocol documentation. This motivates the need for machine reading systems that can interpret the meaning of these natural language instructions, to enhance reproducibility via semantic protocols (e.g. the Aquarium project) and enable robotic automation \cite{doi:10.1021/acssynbio.6b00108} by mapping natural language instructions to executable actions.


\begin{figure}[ht]
\centering
\fbox{\begin{minipage}{0.95\linewidth}
\small
\textbf{Isolation of temperate phages by plaque agar overlay}

1. Melt soft agar overlay tubes in boiling water and place in the 47\textdegree\ C water bath.

2. Remove one tube of soft agar from the water bath. 

3. Add 1.0 mL host culture and either 1.0 or 0.1 mL viral concentrate.

4. Mix the contents of the tube well by rolling back and forth between two hands, and immediately empty the tube contents onto an agar plate. 

5. Sit RT for 5 min.

6. Gently spread the top agar over the agar surface by sliding the plate on the bench surface using a circular motion. 

7. Harden the top agar by not disturbing the plates for 30 min. 

8. Incubate the plates (top agar side down) overnight to 48 h. 

9. Temperate phage plaques will appear as turbid or cloudy plaques, whereas purely lytic phage will appear as sharply defined, clear plaques.
\end{minipage}

}

\caption{An example wet lab protocol. The first seven steps are imperative sentences, and the last sentence describes the end results and their subsequent utilization.}
\label{fig:protocol}
\end{figure}
In this study we take a first step towards this goal by annotating a database of wet lab protocols with semantic actions and their arguments; and conducting initial experiments to demonstrate its utility for machine learning approaches to shallow semantic parsing of natural language instructions. To the best of our knowledge, this is the first annotated corpus of natural language instructions in the biomedical domain that is large enough to enable machine learning approaches. 

There have been many recent data collection and annotation efforts that have initiated natural language processing research in new directions, for example political framing \cite{card2015media}, question answering \cite{rajpurkar2016squad} and cooking recipes \cite{jermsurawong2015predicting}. Although mapping natural language instructions to machine readable representations is an important direction with many practical applications, we believe current research in this area is hampered by the lack of available annotated corpora. Our annotated corpus of wet lab protocols could enable further research on interpreting natural language instructions, with practical applications in biology and life sciences.

\begin{figure*}
   \centerline{ \includegraphics[width=0.99\linewidth, scale=0.4]{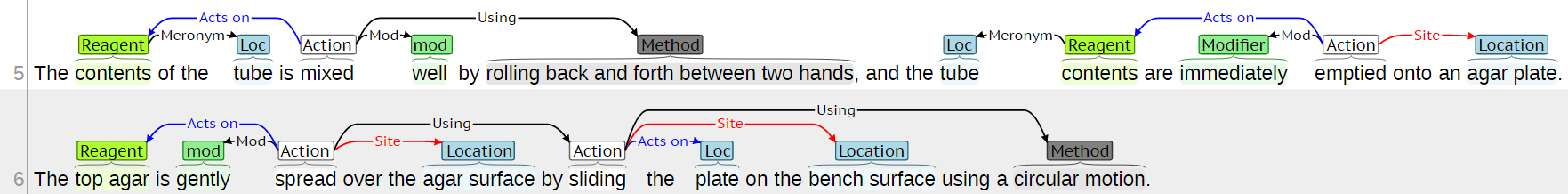}}
 \caption{Example sentences (\#5 and \#6) from the lab protocol in Figure \ref{fig:protocol} as shown in the BRAT annotation interface.}
 
 \label{fig:sent}
\end{figure*}

\begin{figure}[ht]
 
  \centering
    \includegraphics[width=7cm]{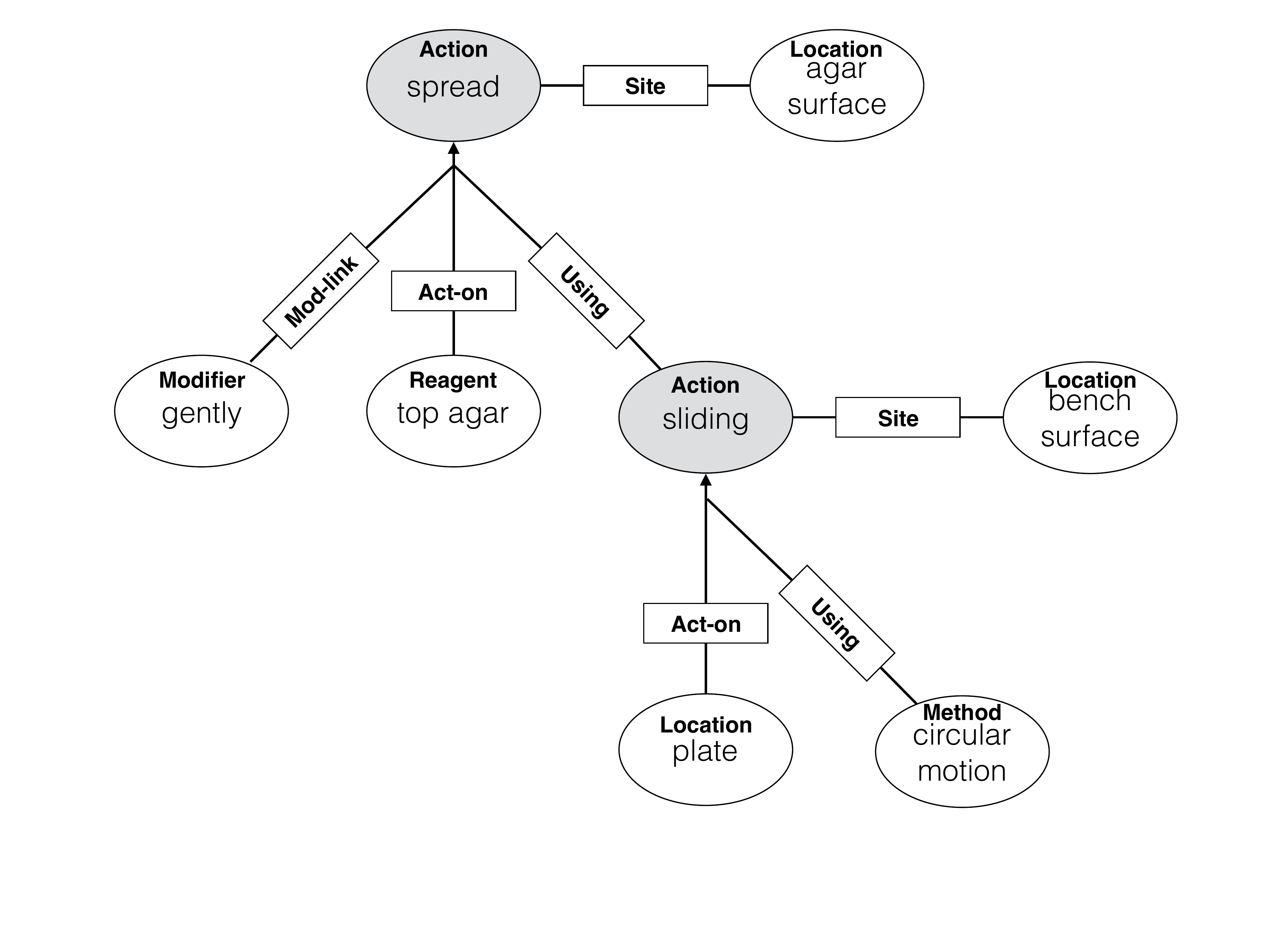}
 \caption{An action graph can be directly derived from annotations as seen in Figure \ref{fig:sent} (example sentence \#6) .}
 \label{fig:graph}
\end{figure}

Prior work has explored the problem of learning to map natural language instructions to actions, often learning through indirect supervision to address the lack of labeled data in instructional domains.  This is done, for example, by interacting with the environment \cite{branavan2009reinforcement,branavan2010reading} or observing weakly aligned sequences of instructions and corresponding actions \cite{chen2011learning,artzi2013weakly}.
In contrast, we present the first steps towards a pragmatic approach based on linguistic annotation (Figure \ref{fig:graph}). We describe our effort to exhaustively annotate wet lab protocols with actions corresponding to lab procedures and their attributes including materials, instruments and devices used to perform specific actions.  As we demonstrate in \S \ref{results}, our corpus can be used to train machine learning models which are capable of automatically annotating lab-protocols with action predicates and their arguments \cite{gildea2002automatic,das2014frame}; this could provide a useful linguistic representation for robotic automation \cite{bollini2013interpreting} and other downstream applications. 


\section{Wet Lab Protocols}

Wet laboratories are laboratories for conducting biology and chemistry experiments which involve chemicals, drugs, or other materials in liquid solutions or volatile phases. Figure \ref{fig:protocol} shows one representative wet lab protocol. Research groups around the world curate their own repositories of protocols, each adapted from a canonical source and typically published in the Materials and Method section at the end of a scientific article in biology and chemistry fields. Only recently has there been an effort to gather collections of these protocols and make them easily available. Leveraging an openly accessible repository of protocols curated on the \href{https://www.protocols.io}{\textit{https://www.protocols.io}} platform, we annotated hundreds of academic and commercial protocols maintained by many of the leading bio-science laboratory groups, including Verve Net, Innovative Genomics Institute and New England Biolabs. The protocols cover a large spectrum of experimental biology, including neurology, epigenetics, metabolomics, cancer and stem cell biology, etc (Table \ref{tab:p_types_stats}). Wet lab protocols consist of a sequence of steps, mostly composed of imperative statements meant to describe an action. They also can contain declarative sentences describing the results of a previous action, in addition to general guidelines or warnings about the materials being used.

\section{Annotation Scheme}

\begin{table*}[]
\centering
\small
\begin{tabular}{|l||c|c|c|c|c|c|}
\hline
Protocol Category & Count & avg \#Sentences & avg \#Words     & avg \#Entities  & avg \#Relations & avg \#Actions   \\
\hline
molecular biology & 186   & 27.42   & 338.06 & 85.25  & 84.20   & 35.77 \\
microbiology      & 105   & 22.07 & 328.94  & 74.46  & 71.71  & 27.89 \\
cell biology      & 94    & 19.23  & 236.74 & 61.09  & 60.95   & 23.93 \\
Plant biology     & 48    & 17.17 & 219.96 & 44.67 & 43.85 & 20.44            \\
Immunology        & 79    & 25.92 & 339.58  & 83.17  & 78.24  & 32.68  \\
chemical biology  & 110   & 14.37 & 188.30  & 46.40   & 47.45 & 19.01 \\
\hline
\end{tabular}
\caption{Statistics of our Wet Lab Protocol Corpus by protocol category.}
    \label{tab:p_types_stats}
\end{table*}

\begin{table}[ht!]
\small
\begin{tabular}{|l||c|c|c|}
 \hline
  & Total & per Protocol & per Sentence\\
 \hline
        \# of sentences & 13679  & 21.99     & --\\
        \# of words     & 177770 & 285.80    & 12.99\\
        \hline
        \# of entities  & 43236 & 69.51 & 3.16\\
        \# of relations & 42425 & 68.21 & 3.10\\
        \# of actions   & 17485 & 28.11 & 1.28\\
 \hline
\end{tabular}
\caption{Statistics of the Wet Lab Protocol Corpus.}
    \label{tab:corpus_stats}
    \vspace{.16in}
\end{table}

In developing our annotation guidelines we had three primary goals: (1) We aim to produce a semantic representation that is well motivated from a biomedical and linguistic perspective; (2) The guidelines should be easily understood by annotators with or without biology background, as evaluated in Table \ref{tab:comparison}; (3) The resulting corpus should be useful for training machine learning models to automatically extract experimental actions for downstream applications, as evaluated in \S \ref{results}.

We utilized the EXACT2 framework \cite{soldatova2014exact2} as a basis for our annotation scheme. We borrowed and renamed 9 object-based entities from EXACT2, in addition, we created 5 measure-based 
({\sc Numerical}, {\sc Generic-Measure}, {\sc Size}, {\sc pH}, {\sc Measure-Type}) and 3 other ({\sc Mention}, {\sc Modifier},  {\sc Seal}) 
entity types. EXACT2 connects the entities directly to the action without describing the type of relations, whereas we defined and annotated 12 types of relations between actions and entities, or pairs of entities (see Appendix for a full description).



For each protocol, the annotators were requested to identify and mark every span of text that corresponds to one of 17 types of entities or an action (see examples in Figure \ref{fig:sent}). Intersection or overlap of text spans, and the subdivision of words between two spans were not allowed. The annotation guideline was designed to keep the span short for entities, with the average length being 1.6 words. For example,  {\sc Concentration} tags are often very short: \textit{60\%} \textit{10x}, \textit{10M}, \textit{1 g/ml}. The  {\sc Method} tag has the longest average span of 2.232 words with examples such as \textit{rolling back and forth between two hands}. The methods in wet lab protocols tend to be descriptive, which pose distinct challenges from existing named entity extraction research in the medical \cite{doi:10.1093/bioinformatics/btg1023} and other domains. After all entities were labelled, the annotators connected pairs of spans within each sentence by using one of 12 directed links to capture various relationships between spans tagged in the protocol text. While most protocols are written in scientific language, we also observe some non-standard usage, for example using \textit{RT} to refer to \textit{room temperature}, which is tagged as {\sc Temperature}. 




\section{Annotation Process}

\begin{table}[ht!]
    \centering
    \small
    \begin{tabular}{ |l||c|c|  }
        \hline
        Annotators          & Entities+Actions  & Relations \\
        \hline
        Biologist-Linguist  & 0.7600    & 0.6084   \\
        Biologist-Other     & 0.7621    & 0.6619  \\
        Linguist-Other      & 0.7574    & 0.6753  \\
        \hline
        \hline
        all 4 coders        & 0.7599      & 0.6625    \\
 \hline
\end{tabular}
\caption{ Inter-annotator agreement (Krippendorff's $\alpha$) between annotators with biology, linguistics and other backgrounds.}
    \label{tab:comparison}
\end{table}

Our final corpus consists of 622 protocols annotated by a team of 10 annotators. Corpus statistics are provided in Table \ref{tab:p_types_stats} and \ref{tab:corpus_stats}. In the first phase of annotation, we worked with a subset of 4 annotators including one linguist and one biologist to develop the annotation guideline for 6 iterations. For each iteration, we asked all 4 annotators to annotate the same 10 protocols and measured their inter-annotator agreement, which in turn helped in determining the validity of the refined guidelines. The average time to annotate a single protocol of 40 sentences was approximately 33 minutes, across all annotators.

\subsection{Inter-Annotator Agreement}
We used Krippendorff's $\alpha$ for nominal data \cite{krippendorff2004content} to measure the inter-rater agreement for entities, actions and relations. For entities, we measured agreement at the word-level by tagging each word in a span with the span's label.  To evaluate inter-rater agreement for relations between annotated spans, we consider every pair of spans within a step and then test for matches between annotators (partial entity matches are allowed).   We then compute Krippendorff's $\alpha$ over relations between matching pairs of spans.  Inter-rater agreement for entities, actions and relations is presented in Figure \ref{tab:comparison}.



\section{Methods}
To demonstrate the utility of our annotated corpus, we explore two machine learning approaches for extracting actions and entities: a maximum entropy model and a neural network tagging model. We also present experiments for relation classification. We use the standard precision, recall and F$_1$ metrics to evaluate and compare the performance.

\subsection{Maximum Entropy (MaxEnt) Tagger}
In the maximum entropy model for action and entity extraction \cite{borthwick1999maximum}, we used three types of features based on the current word and context words within a window of size 2: 
\begin{itemize}
    \vspace{-.1in}
    \item \textbf{Parts of speech features} which were generated by the GENIA POS Tagger \cite{tsuruoka2005bidirectional}, which is specifically tuned for biomedical texts;
    \vspace{-.1in}
    \item \textbf{Lexical features} which include unigrams, bigrams as well as their lemmas and synonyms from WordNet \cite{miller1995wordnet} are used;
    \vspace{-.1in}
    \item \textbf{Dependency parse features} which include dependent and governor words as well as the dependency type to capture syntactic information related to actions, entities and their contexts. We used the Stanford dependency parser \cite{chen2014fast}. 
\end{itemize}

\subsection{Neural Sequence Tagger}
We utilized the state-of-the-art Bidirectional LSTM with a Conditional Random Fields (CRF) layer \cite{ma2016end,lample2016neural,plank2016multilingual}, initialized with 200-dimentional word vectors
pretrained on 5.5 billion words from PubMed and PMC biomedical texts  \cite{moen2013distributional}. Words unseen in the pretrained vocabulary were randomly initialized using a uniform distribution in the range (-0.01, 0.01). We used Adadelta \cite{zeiler2012adadelta} optimization with a mini-batch of 16 sentences and trained each network with 5 different random seeds, in order to avoid any outlier results due to randomness in the model initialization.

\subsection{Relation Classification}
To demonstrate the utility of the relation annotations, we also experimented with a maximum entropy model for relation classification using features shown to be effective in prior work \cite{li2014incremental,guodong2005exploring,kambhatla2004combining}. The features are divided into five groups:
\begin{itemize}
    \vspace{-.1in}
    \item \textbf{Word features} which include the words contained in both arguments, all words in between, and context words surrounding the arguments;
    \vspace{-.1in}
     \item \textbf{Entity type features} which include action and entity types associated with both arguments; 
     \vspace{-.1in}
     \item \textbf{Overlapping features} which are the number of words, as well as actions or entities, in between the candidate entity pair; 
     \vspace{-.1in}
     \item \textbf{Chunk features} which are the chunk tags of both arguments predicted by the GENIA tagger; 
     \vspace{-.1in}
      \item \textbf{Dependency features} which are context words related to the arguments in the dependency tree according to the Stanford Dependency Parser. 
\end{itemize}
Also included are features indicating whether the two spans are in the same noun phrase, prepositional phrase, or verb phrase.

\begin{table}
\small
\centering
\begin{tabular}{|l||c||c|c|}

 \hline
 \thead{Entity/Action\\(freq. in test set)} & \thead{MaxEnt} & \thead{BiLSTM} & \thead{BiLSTM\\+ CRF} \\
 \hline
Action \tiny(3519)          & 83.87           & 85.95 & 86.89  \\
\hline
Amount \tiny(886)           & 68.25      & 81.59 & 82.34            \\
Conc. \tiny(273)    & 56.84      & 65.36 & 76.36       \\
Device \tiny(408)           & 49.14      & 58.73 & 64.02        \\
Gen.-Measure \tiny(91)   & 05.88      & 06.45 & 25.68     \\
Location \tiny(1007)        & 61.07      & 69.57 & 73.53      \\
Meas.-Type \tiny(50)      & 15.38      & 18.75 & 21.62     \\
Mention \tiny(37)           & 43.37      & 52.31 & 57.97     \\
Method  \tiny(177)          & 37.97      & 30.60 & 38.21   \\
Modifier \tiny(720)         & 50.86      & 56.90 & 59.34 \\
Numerical \tiny(129)        & 39.70      & 47.84 & 49.80 \\
Reagent \tiny(2486)         & 60.54      & 71.34 & 74.55 \\
Seal \tiny(43)              & 49.52      & 54.05 & 66.67 \\
Size \tiny(69)              & 19.35      & 24.82 & 26.92      \\
Speed  \tiny(200)           & 74.88      & 85.31 & 91.00      \\
Temperature \tiny(469)      & 80.69      & 86.68 & 91.90      \\
Time \tiny(708)             & 83.68      & 92.69 & 93.94     \\
pH \tiny(21)                & 41.86      & 53.66 & 70.00     \\
\hline
\hline
Macro-avg F1      & 49.23      & 58.81 & 64.44 \\
Micro-avg F1      & 68.03      & 74.99 & 78.03   \\

 \hline
\end{tabular}

\caption{ F1 scores for segmenting and classifying entities and action triggers compared across the various models.}
    \label{tab:individual_results}
\end{table}

\begin{table}[t]
\small
\begin{center}
\begin{tabular}{ |l||c|c|c|c|c|c|c|  }
 \hline
    MaxEnt Model                 &  \multicolumn{3}{c|}{Relations} \\
\hline
    Features                    & P     & R     & F1    \\
\hline

Words                           & 66.16 & 46.84 & 54.85 \\
+ Entity Type                   & 78.93 & 72.75 & 75.72 \\
+ Overlap                       & 80.81 & 74.73 & 77.65 \\
+ Base Phrase Chunking          & 81.04 & 76.52 & 78.71 \\
+ Dependency Tree               & 80.98 & 77.04 & 78.96 \\

 \hline
\end{tabular}
\end{center}
\caption{Precision, Recall and F1 (micro-average) of the maximum entropy model for relation classification, as each feature is added.}
    \label{table:maxent_relations}
\end{table}

\begin{table*}[ht]
\small
\begin{center}
\begin{tabular}{ |l||c|c|c|c|c|c|c|  }
 \hline
    MaxEnt Model                 &  \multicolumn{3}{c|}{Actions} & \multicolumn{3}{c|}{Entities} \\
\hline
    Features        & P     & R     & F1       & P      & R      & F1     \\
\hline

POS                     & 74.83 & 79.94 & 77.30*     & 26.66  & 27.93  & 28.77  \\
uni/bigram                  & 76.29 & 69.59 & 72.79     & 43.75  & 32.93  & 37.58 \\
POS, uni/bigram              & 79.77 & 85.51 & 82.54     & 49.83  & 54.51  & 52.07 \\
POS, uni/bigram, lem./syn.   & 80.10 & 85.56 & 82.74     & 49.79  &54.54   & 52.06  \\
POS, uni/bigram, lem./syn., dep.            & \bf{81.65} & \bf{86.22} & \bf{83.87}     &  \bf{57.04}  &  \bf{63.03}  & \bf{59.90}*  \\

 \hline
\end{tabular}
\end{center}
\caption{Performance of  maximum entropy model with various features.*The POS features are especially useful for recognizing actions; dependency based features are more helpful for entities than actions.}
    \label{table:maxent}
\end{table*}

\begin{table}[htp]
    \centering
    \small
    \begin{tabular}{|l|l|}
        \hline
        POS tag {\tiny(freq.)}& Top 3 examples \\
        \hline
        VB {\tiny(9345)}  & Add{\tiny(1404)}, Incubate{\tiny(638)}, Remove{\tiny(396)} \\
        VBG {\tiny(755)}   & adding{\tiny(112)}, inverting{\tiny(89)}, pipetting{\tiny(34)} \\
        VBN {\tiny(727)}   & added{\tiny(43)}, stored{\tiny(38)}, incubated{\tiny(38)} \\
        VBP {\tiny(512)}   & Do{\tiny(80)}, mix{\tiny(38)}, pour{\tiny(33)} \\
        VBD {\tiny(147)}   & resuspend{\tiny(25)}, put{\tiny(20)}, kept{\tiny(8)}  \\
        VBZ {\tiny(44)}    & remains{\tiny(5)}, covers{\tiny(4)}, washes{\tiny(3)} \\
        NN  {\tiny(4248)}  & Centrifuge{\tiny(324)}, Transfer{\tiny(301)}, Place{\tiny(215)} \\
        NNP {\tiny(1551)}  & Mix{\tiny(335)}, Wash{\tiny(277)}, Vortex{\tiny(114)} \\
        NNS {\tiny(80)}    & washes{\tiny(9)}, to{\tiny(7)}, dilutions{\tiny(4)} \\
        JJ  {\tiny(576)}   & dry{\tiny(66)}, Apply{\tiny(26)}, decant{\tiny(23)} \\
        OTHER {\tiny(1080)}   & not{\tiny(111)}, off{\tiny(110)}, up{\tiny(105)} \\
        \hline
    \end{tabular}
    \caption{ Frequency of different part-of-speech (POS) tags for action words. Majority of the action words either fall under the verb POS tags (VBs 60.48\%) or nouns (NNs 30.84\%). The GENIA POS tagger is under-identifying verbs in the wet lab protocols, tagging some as adjectives (JJ).}
    \label{tab:pos_freq}
\end{table}

\section{Results}
\label{results}
The full annotated dataset of 622 protocols are randomly split into training, dev and test sets using a 6:2:2 ratio. The training set contains 374 protocols of 8207 sentences, development set contains 123 protocols of 2736 sentences, and test set contains 125 protocols of 2736 sentences. We use the evaluation script from the CoNLL-03 shared task \cite{tjong2003introduction}, which requires exact matches of label spans and does not reward partial matches. During the data preprocessing, all digits were replaced by `0'.

\subsection{Entity Identification and Classification}

Table \ref{tab:individual_results} shows the performance of various methods for entity tagging. We found that the BiLSTM-CRF model consistently outperforms other methods, achieving an overall F1 score of 86.89 at identifying action triggers and 72.61 at identifying and classifying entities.

Table \ref{table:maxent} shows the system performance of the MaxEnt tagger using various features. Dependency based features have the highest impact on the detection of entities, as illustrated by the absolute drop of  7.84\% in F-score when removed. Parts of speech features alone are the most effective in capturing action words. This is largely  due to action words appearing as verbs or nouns in the majority of the sentences as shown in Table \ref{tab:pos_freq}. We also notice that the GENIA POS tagger, which is is trained on Wall Street Journal and biomedical abstracts in the GENIA and PennBioIE corpora, under-identifies verbs in wet lab protocols. We suspect this is due to fewer imperative sentences in the training data. We leave further investigation for future work, and hope the release of our dataset can help draw more attention to NLP research on instructional languages. 

\subsection{Relation Classification}
Finally, precision and recall at relation extraction are presented in Table 5. We used gold action and entity segments for the purposes of this particular evaluation. We obtained the best performance when using all feature sets.

\section{Conclusions}
In this paper, we described our effort to annotate wet lab protocols with actions and their semantic arguments. We presented an annotation scheme that is both biologically and linguistically motivated and demonstrated that non-experts can effectively annotate lab protocols. Additionally, we empirically demonstrated the utility of our corpus for developing machine learning approaches to shallow semantic parsing of instructions.  Our annotated corpus of protocols is available for use by the research community.

\section*{Acknowledgement}
We would like to thank the annotators: Bethany Toma, Esko Kautto, Sanaya Shroff, Alex Jacobs, Berkay Kaplan, Colins Sullivan, Junfa Zhu, Neena Baliga and Vardaan Gangal. We would like to thank Marie-Catherine de Marneffe and anonymous reviewers for their feedback.  


\bibliography{naaclhlt2018}
\bibliographystyle{acl_natbib}

\clearpage
\appendix
\section{Annotation Guidelines}
The wet lab protocol dataset annotation guidelines were designed primarily to provide a simple description of the various actions and their arguments in protocols so that it could be more accessible and be effectively used by non-biologists who may want to use this dataset for various natural language processing tasks such as action trigger detection or relation extraction. In the following sub-sections we summarize the guidelines that were used in annotating the 622 protocols as we explore the actions, entities and relations that were chosen to be labelled in this dataset. 

\subsection{Actions}

Under a broad categorization, Action is a process of doing something, typically to achieve an aim. In the context of wet lab protocols, action mentions in a sentence or a step are deliberate but short descriptions of a task tying together various entities in a meaningful way.  Some examples of action words, (categorized using GENIA POS tagger), are present in Table \ref{tab:pos_freq} along with their frequencies.

\subsection{Entities}
\begin{figure}[h]
    \centering
    \includegraphics[width=0.9\linewidth]{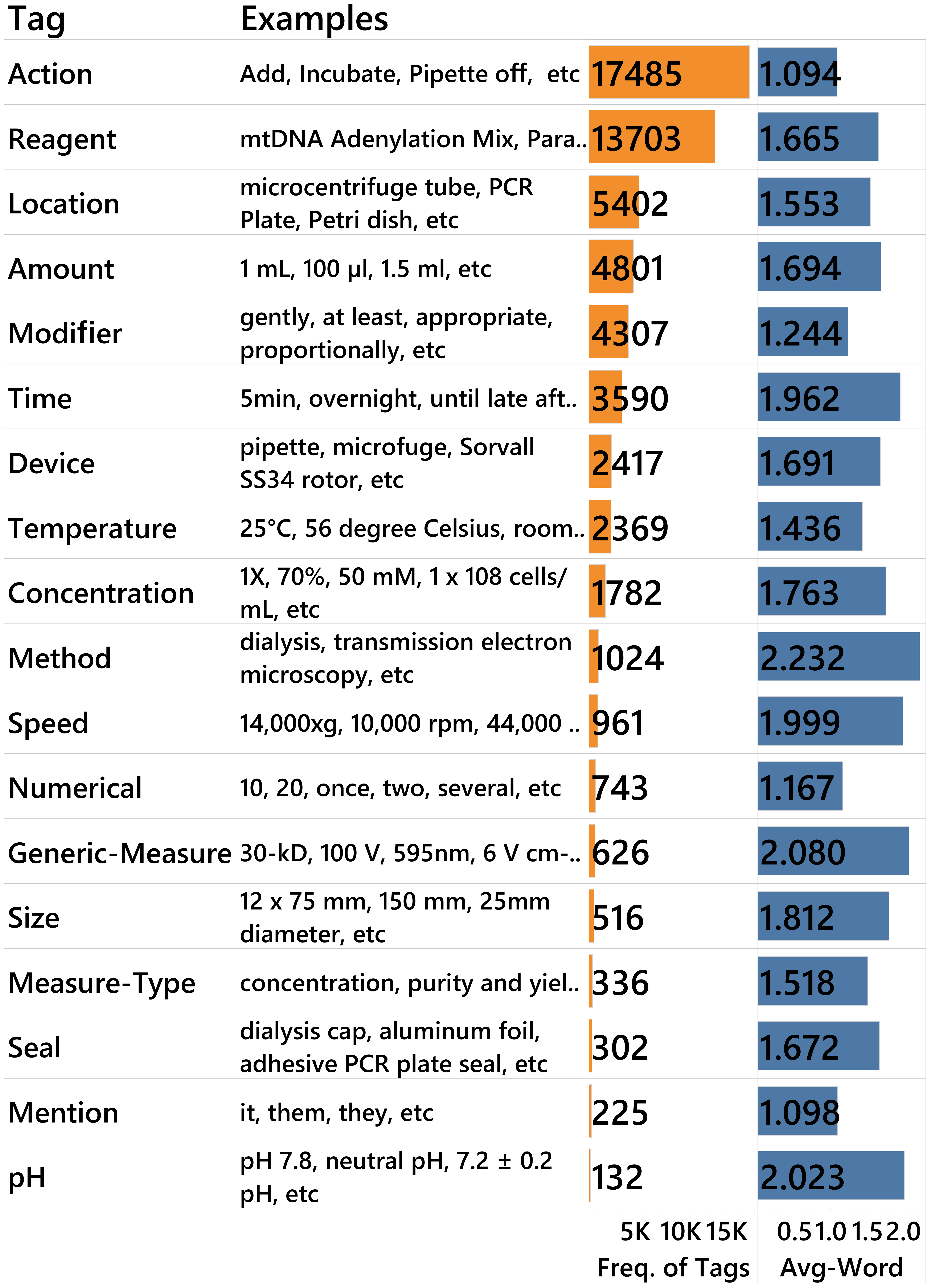}
    \caption{ Examples, Frequency and Avg-Word for actions and entities.}
    \label{fig:freq_v_words}
\end{figure}

We broadly classify entities commonly seen in protocols under 17 tags. Each of the entity tags were designed to encourage short span length, with the average number of words per entity tag being $1.6$. For example, \verb!Concentration! tags are often very short: \textit{60\%} \textit{10x}, \textit{10M}, \textit{1 g/ml}, while the \verb!Method! tag has the longest average span of $2.232$ words with examples such as \textit{rolling back and forth between two hands} (as seen in Figure \ref{fig:freq_v_words}). The methods in wet lab protocols tend to be descriptive, which pose distinct challenges from existing named entity extraction research in the medical and other domains.

\subsubsection{Object Based Entities}

\noindent
\textbf{Reagent:} A substance or mixture for use in any kind of reaction in preparing a product because of its chemical or biological activity.

\noindent
\textbf{Location:} Containers for reagents or other physical entities. They lack any operation capabilities other than acting as a container. These could be laboratory glassware or plastic tubing meant to hold chemicals or biological substances. 

\noindent
\textbf{Device:} A machine capable of acting as a container as well as performing a specific task on the objects that it holds. A device and a location are similar in all aspects except that a device performs a specific set of operations on its contents, usually illustrated in the sentence itself, or sometimes implied. 

\noindent
\textbf{Seal:} Any kind of lid or enclosure for the location or device. It could be a cap, or a membrane that actively participates in the protocol action, and hence is essential to capture this type of entity.

\begin{table*}[]
\centering
\begin{tabular}{|m{1.8cm}|m{7.9cm}|m{5cm}|}
    \hline
    Label & Syntax/Rules & Example \\
    \hline
    Acts-on & Action $\Rightarrow$ Reagent $\vert$ Location $\vert$ Mention $\vert$ Device $\vert$ Seal & \vspace{0.12cm}\includegraphics[scale=0.7]{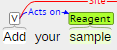}\\
    \hline
    Creates & Action $\Rightarrow$ Reagent $\vert$ Mention & \vspace{0.12cm}\includegraphics[scale=0.7]{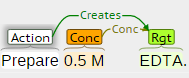} \\
    \hline
    Site & Action $\Rightarrow$ Location $\vert$ Device $\vert$ Mention $\vert$ Reagent
    & \vspace{0.12cm}\includegraphics[scale=0.67]{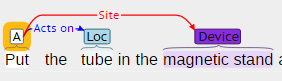} \\
    \hline 
    Using & Action $\Rightarrow$ Method $\vert$ Action $\vert$ Seal $\vert$ Device $\vert$ Mention $\vert$ Reagent $\vert$ Location
    & \vspace{0.12cm}\includegraphics[scale=0.7]{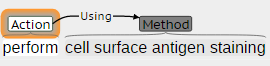}\\
    \hline
    Setting & Action $\vert$ Device $\vert$ Modifier $\Rightarrow$ Method $\vert$ Action $\vert$ Seal $\vert$ Device $\vert$ Mention $\vert$ Reagent $\vert$ Location
    & \vspace{0.12cm}\includegraphics[scale=0.66]{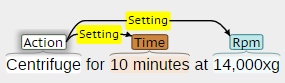} \\
    \hline
    Count & Action $\Rightarrow$ Numerical 
    & \vspace{0.12cm}\includegraphics[scale=0.7]{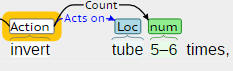} \\
    \hline
    Measure-Type-Link & Action $\Rightarrow$ Measure-Type 
    & \vspace{0.12cm}\includegraphics[scale=0.6]{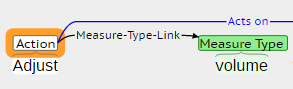} \\
    \hline
    Coreference & Mention $\Rightarrow$ [Every other entity]
    & \vspace{0.12cm}\includegraphics[scale=0.7]{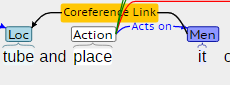} \\
    \hline
    Mod-Link & [Every Entity or Action] $\Rightarrow$ Modifier
    & \vspace{0.12cm}\includegraphics[scale=0.7]{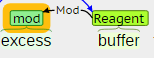} \\
    \hline
    Measure & Reagent $\vert$ Location $\vert$ Device $\vert$ Mention $\vert$ Seal $\Rightarrow$ Amount $\vert$ Numerical $\vert$ Size $\vert$ Concentration $\vert$ Generic-Measure $\vert$ pH
    & \vspace{0.12cm}\includegraphics[scale=0.7]{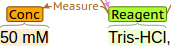} \\
    \hline
    Meronym & Reagent $\vert$ Location $\vert$ Device $\vert$ Mention $\vert$ Seal $\Rightarrow$ Reagent $\vert$ Location $\vert$ Device $\vert$ Mention $\vert$ Seal
    & \vspace{0.12cm}\includegraphics[scale=0.7]{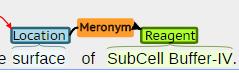} \\
    \hline
    Or & [All Entities or Action] $\Rightarrow$ [All Entities or Action]
    & \vspace{0.12cm}\includegraphics[scale=0.7]{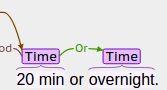} \\
    \hline
    Of-Type & Generic-Measure $\vert$ Numerical $\Rightarrow$ Measure-Type
    & \vspace{0.12cm}\includegraphics[scale=0.5]{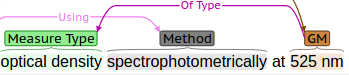} \\
    \hline
    
\end{tabular}
\caption{Relations along with their rules and examples}
\end{table*}

\subsubsection{Measure Based Entities}

\noindent
\textbf{Amount:} The amount of any reagent being used in a given step, in terms of weight or volume. 

\noindent
\textbf{Concentration:} Measure of the relative proportions of two or more quantities in a mixture. Usually in terms of their percentages by weight or volume.

\noindent
\textbf{Time:} Duration of a specific action described in a single step or steps, typically in secs, min, days, or weeks.

\noindent
\textbf{Temperature:} Any temperature mentioned in degree Celsius, Fahrenheit, or Kelvin. 

\noindent
\textbf{Method:} A word or phrase used to concisely define the procedure to be performed in association with the chosen action verb. It’s usually a noun, but could also be a passive verb. 

\noindent
\textbf{Speed:} Typically a measure that represents rotation per min for centrifuges. 

\noindent
\textbf{Numerical:} A generic tag for a number that doesn't fit time, temp, etc and which isn't accompanied by its unit of measure.

\noindent
\textbf{Generic-Measure:} Any measures that don't fit the list of defined measures in this list.

\noindent
\textbf{Size} A measure of the dimension of an object. For example: length, area or thickness.

\noindent
\textbf{Measure-Type:} A generic tag to mark the type of measurement associated with a number.

\noindent
\textbf{pH:} measure of acidity or alkalinity of a solution. 

\subsubsection{Parts of Speech based Entities}

\noindent
\textbf{Modifier:} A word or a phrase that acts as an additional description of the entity it is modifying. 
For example, \textit{quickly mix} vs \textit{slowly mix} are clearly two different actions, informed by their modifiers "quickly" or "slowly" respectively.

\noindent
\textbf{Mention:} Words that can refer to an object mentioned earlier in the sentence. 

\subsection{Relations}

\subsubsection{Action Relations (Action - Entity)}

\noindent
\textbf{Acts-On:} Links the reagent, or location that the action acts on, typically linking the direct objects in the sentence to the action.

\noindent
\textbf{Creates:} This relation marks the physical entity that the action creates. 

\noindent
\textbf{Site:} A link that associates a Location or Device to an action. It indicates that the Device or Location is the site where the action is performed. It is also used as a way to indicate which entity will finally hold/contain the result of the action.

\noindent
\textbf{Using:} Any entity that the action verb makes ‘use’ of is linked with this relation. 

\noindent
\textbf{Setting:} Any measure type entity that is being used to set a device is linked to the action that is attempting to use that numerical. 

\noindent
\textbf{Count:} A Numerical entity that represents the number of times the action should take place.

\noindent
\textbf{Measure Type Link:} Associates an action to a Measure Type entity that the Action is instructing to measure.

\subsubsection{Binary Relations (Entity - Entity)}

\noindent
\textbf{Coreference:} A link that associates two phrases when those two phrases refer to the same entity. 

\noindent
\textbf{Mod Link:} A Modifier entity is linked to any entity that it is attempting to modify using this relation.

\noindent
\textbf{Settings:} Links devices to their settings directly, only if there is no Action associated with those settings.

\noindent
\textbf{Measure:} A link that associates the various numerical measures to the entity its trying to measure directly.

\noindent
\textbf{Meronym:} Links reagents, locations or devices with materials contained in the reagent, location or device. 

\noindent
\textbf{Or:} Allows chaining multiple entities where either of them can be used for a given link. 

\noindent
\textbf{Of-Type:} used to specify the Measure-Type of a Generic-Measure or a Numerical, if the sentence contains this information.

\end{document}